\documentclass[letterpaper, 10 pt, conference]{ieeeconf}  

\IEEEoverridecommandlockouts                              

\usepackage{cite}
\usepackage{amsmath,amssymb,amsfonts}
\usepackage{algorithmic}
\usepackage{graphicx}
\usepackage{textcomp}
\usepackage{xcolor}

\usepackage{multicol}
\usepackage{wrapfig}
\usepackage{subcaption}
\usepackage{amsmath}
\usepackage[ruled,vlined,linesnumbered]{algorithm2e}

\let\labelindent\relax
\usepackage{enumitem}

\def\BibTeX{{\rm B\kern-.05em{\sc i\kern-.025em b}\kern-.08em
    T\kern-.1667em\lower.7ex\hbox{E}\kern-.125emX}}
\begin{document}

\title{\LARGE \bf
Portfolio of Solving Strategies in CEGAR-based\\Object Packing and Scheduling for Sequential 3D Printing
}

\newtheorem{definition}{Definition}
\newtheorem{proposition}{Proposition}
\newtheorem{corollary}{Corollary}



\author{Pavel Surynek$^{1}$ \thanks{$^{1}$ Faculty of Information Technology, Czech Technical University in Prague, Th\'{a}kurova 9, 160 00 Praha 6, Czechia {\tt\small pavel.surynek@fit.cvut.cz}}}

\maketitle

\begin{abstract}
Computing power that used to be available only in supercomputers decades ago especially their parallelism is currently available in standard personal computer CPUs even in CPUs for mobile telephones. We show how to effectively utilize the computing power of modern multi-core personal computer CPU to solve the complex combinatorial problem of object arrangement and scheduling for sequential 3D printing. We achieved this by parallelizing the existing CEGAR-SEQ algorithm that solves the sequential object arrangement and scheduling by expressing it as a linear arithmetic formula which is then solved by a technique inspired by counterexample guided abstraction refinement (CEGAR). The original CEGAR-SEQ algorithm uses an object arrangement strategy that places objects towards the center of the printing plate. We propose alternative object arrangement strategies such as placing objects towards a corner of the printing plate and scheduling objects according to their height. Our parallelization is done at the high-level where we execute the CEGAR-SEQ algorithm in parallel with a portfolio of object arrangement strategies, an algorithm is called Porfolio-CEGAR-SEQ. Our experimental evaluation indicates that Porfolio-CEGAR-SEQ outperforms the original CEGAR-SEQ. When a batch of objects for multiple printing plates is scheduled, Portfolio-CEGAR-SEQ often uses fewer printing plates than CEGAR-SEQ.

Keywords: 3D printing, FDM 3D printing, Cartesian 3D printer, sequential printing, collision avoidance, rectangle packing, object packing, 3D packing, portfolio solver, object arrangement strategies, strategy portfolio

\end{abstract}


\section{Introduction}

Additive manufacturing, i.e. 3D printing, is an increasingly important alternative to traditional manufacturing processes. At the same time, 3D printing, due to its nature, is very close to robotics and to the techniques of artificial intelligence and combinatorial optimization that are used in robotics, since the 3D printer itself can be viewed as a special robot \cite{DBLP:journals/cad/GaoZRRCWWSZZ15}.

A standard Cartesian fused deposition modeling (FDM) 3D printer creates objects on a rectangular printing plate (usually heated) by gradually drawing individual slices of printed objects, where these slices are very thin, approximately tenths of a millimeter. Printing is performed using a print head with an extruder, which applies material through a narrow nozzle. The movement of the print head is ensured by a printer mechanism that allows the head to move in all $x$, $y$, and $z$ coordinates.

One of the important tasks of combinatorial optimization in 3D printing is the arrangement of printed objects on the printing plate so that the space of the place is used effectively \cite{multi-objective-packing2014,DBLP:conf/ki/EdelkampW15,DBLP:journals/cg/CaoTPZL21}.

\begin{figure*}[t]
    \centering
    \begin{subfigure}{0.33\textwidth}
       \includegraphics[trim={0.5cm 0.5cm 0.5cm 0.5cm},clip,width=1.0\textwidth]{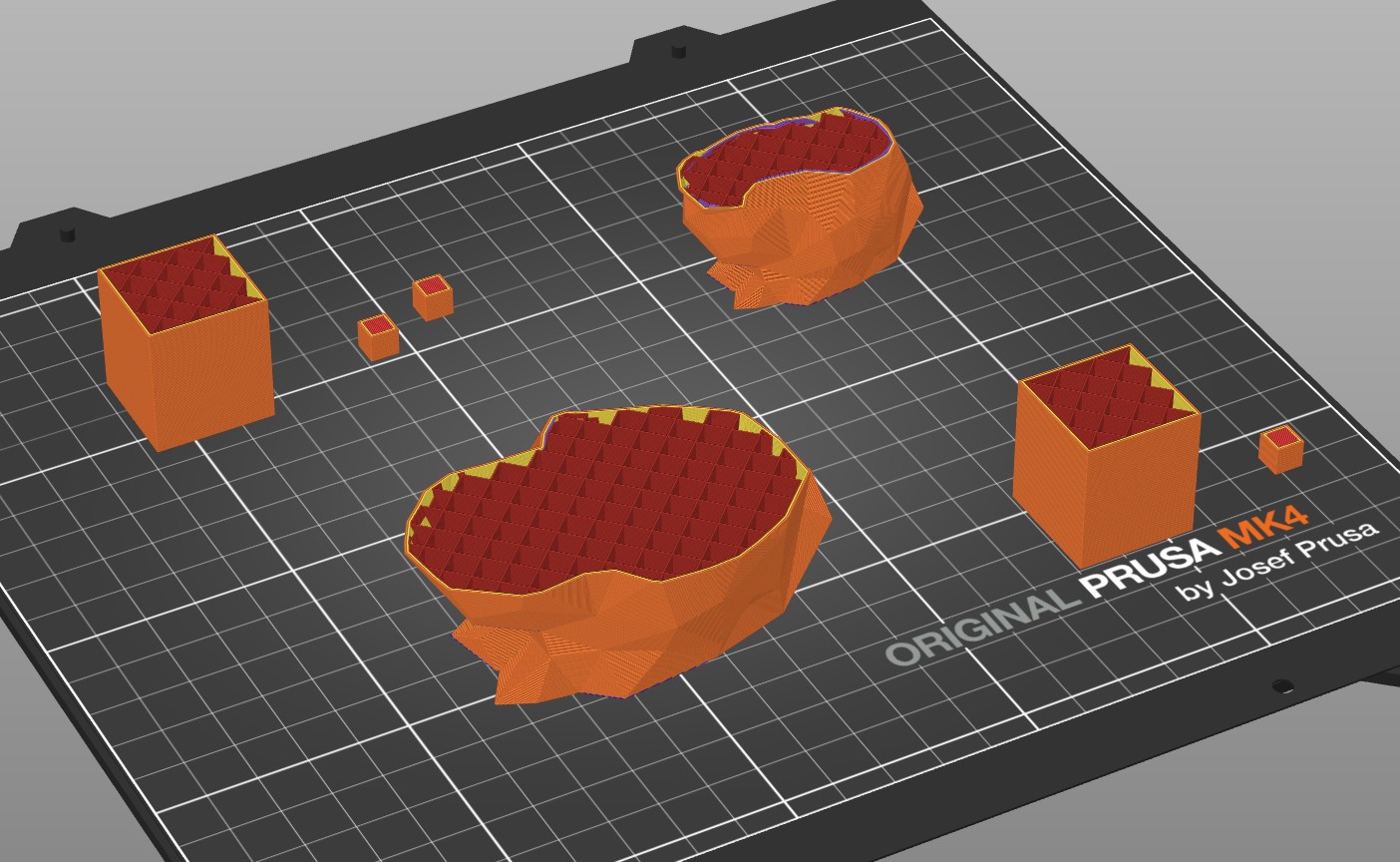}
    \end{subfigure}
    \begin{subfigure}{0.31\textwidth}
       \includegraphics[trim={0.5cm 0.5cm 0.5cm 0.5cm},clip,width=1.0\textwidth]{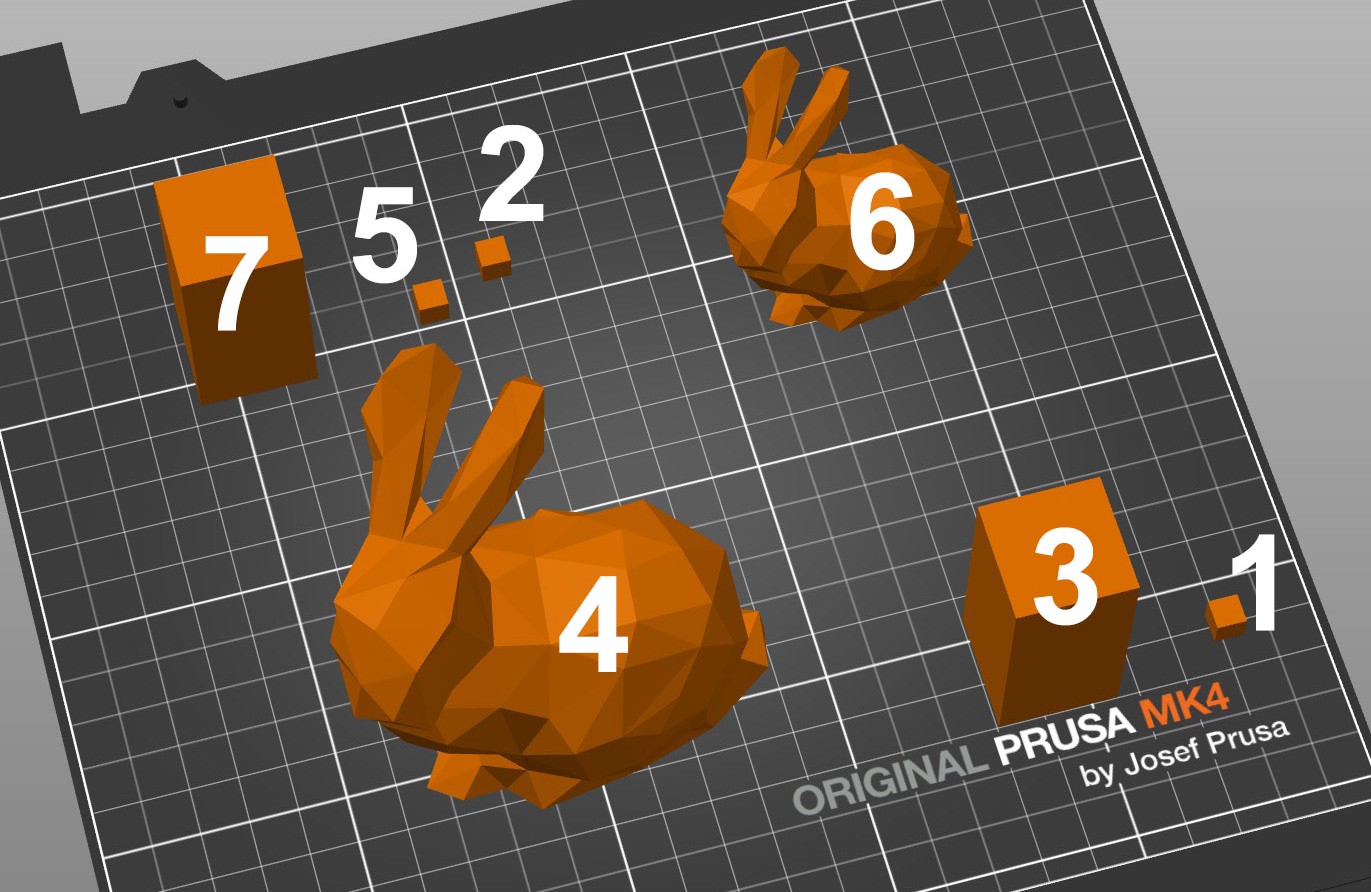}
    \end{subfigure}
    \begin{subfigure}{0.345\textwidth}
       \includegraphics[trim={0.5cm 0.5cm 0.5cm 0.5cm},clip,width=1.0\textwidth]{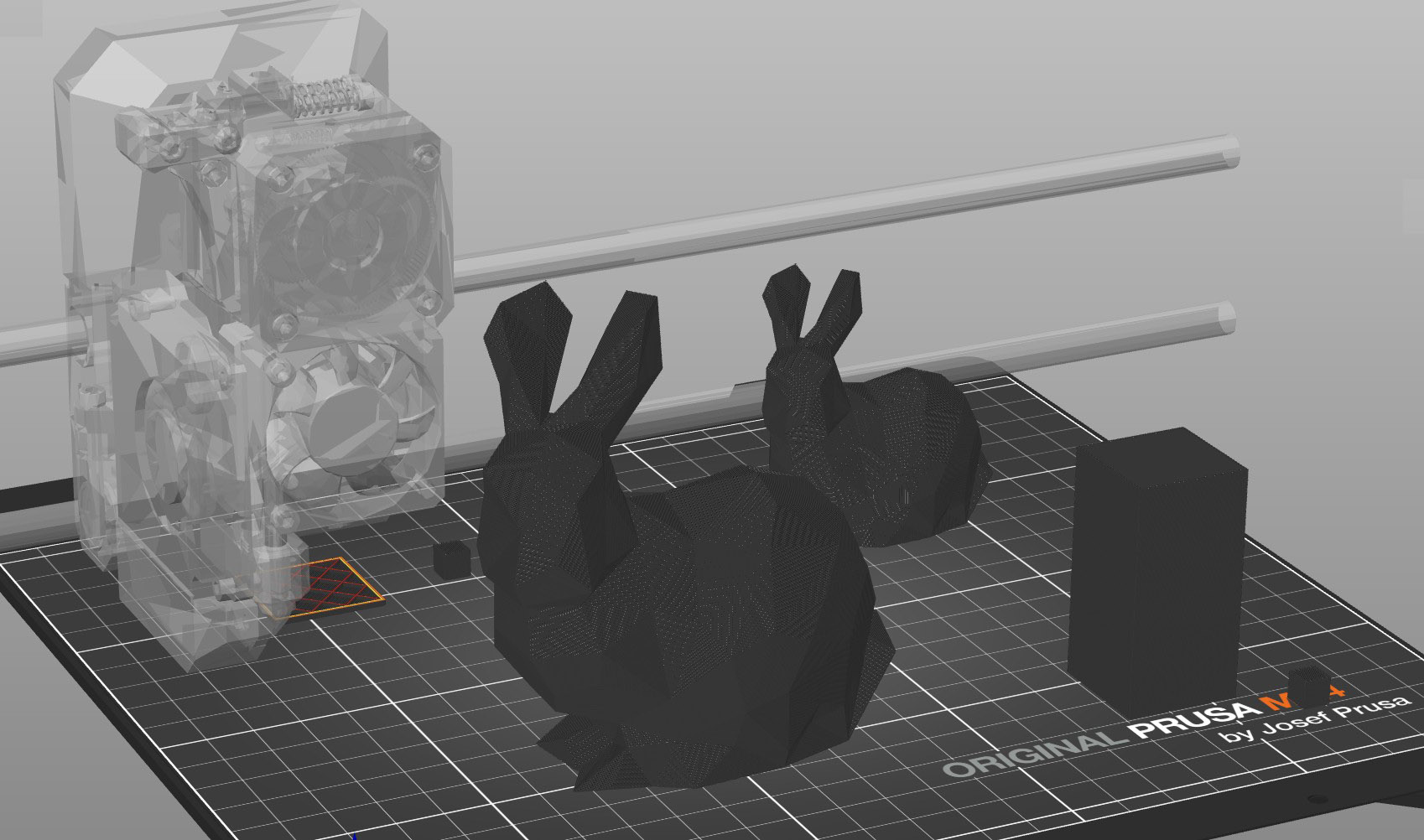}
    \end{subfigure}
    \caption{Standard 3D printing slice by slice and sequential 3D printing where objects are completed one by one shown in Prusa Slicer \cite{prusa-slicer-2025}. The ordering of objects for sequential printing is shown by numbers. Printer extruder and gantry must avoid previously printed objects in the sequential case (printing of the last object is shown). The sequential arrangement and schedule has been obtained by the CEGAR-SEQ algorithm \cite{DBLP:conf/iros/SurynekBMK25}.}
    \vspace{-0.5cm}    
    \label{fig:seq-print}
\end{figure*}

In this work, we deal with the recently introduced task of {\em sequential 3D printing} \cite{DBLP:conf/case/SurynekBMK25,DBLP:conf/iros/SurynekBMK25}, where not print all objects are printed slice-by-slice at once, but individual objects are completed one after other, while individual objects still being printed in the standard slice-by-slice manner. As noted in \cite{DBLP:conf/iros/SurynekBMK25}, sequential printing is particularly challenging because the objects need to be arranged on the printing plate in such a way that the print head and other mechanical parts, such as the gantry on which the print head is mounted, avoid previously printed objects (see Figure \ref{fig:seq-print} for illustrations). Moreover, sequential printing does not only mean spatial arrangement of objects on a printing plate, but also determining the order in which the objects are printed.

Specifically, we address computational difficulty of the sequential printing problem by utilizing parallelism of modern CPUs. The previous work, namely the CEGAR-SEQ algorithm \cite{DBLP:conf/iros/SurynekBMK25}, translates the problem of sequential printing to a linear arithmetic formula that is subsequently solved by an off-the-shelf solver. This approach has a significant limitation that the solver runs in non-parallel mode and therefore does not fully utilize modern multi-core CPUs. Moreover, low-level parallelization at the level of fine grained operations of the solving algorithm based on search usually does not help much due to its exponential complexity.

Our approach is different, we identify parameters of the original CEGAR-SEQ algorithm that has a significant impact on its answer. Then we run CEGAR-SEQ with different settings of the parameters in parallel and eventually choose the best answer. We call the different setting of the parameters a {\bf portfolio} and the resulting algorithmic framework is called Portfolio-CEGAR-SEQ. The parameters we use to build the portfolio are in fact various heuristic strategies such as specific preference for placement of objects on the printing plate (the original CEGAR-SEQ optimizes placement of objects towards the center of the printing plate) or the ordering that determines selection of objects for a given printing plate in the case of sequential object scheduling for multiple printing plates.


Sequential printing has far-reaching significance for modern 3D printing, it can help to tackle the following challenges:

\begin{enumerate}[label=(\roman*)]
\item increasing the robustness of the printing process to errors (in case of failure, we do not have to repeat the entire print, but only the unfinished objects)
\item elimination of difficulties caused by frequent movements of the print head between objects (such as {\em stringing} etc., can also help in increasing the printing speed)
\item minimize the number of time consuming color changes during multi-color printing, where a classic example is printing each object in a different color.
\end{enumerate}

\subsection{Related Work}

Abstract versions of the problem of object arrangement on a printing plate have been studied in literature and is known as {\em object packing} or {\em object stacking} \cite{DBLP:journals/cg/CaoTPZL21}. Search-based algorithms have been developed for the related problem of {\em rectangle packing} \cite{DBLP:conf/socs/HuangK11,DBLP:journals/jair/HuangK13}. As shown in \cite{DBLP:conf/aips/Korf03}, rectangle packing is NP-hard, which means that object arrangement and scheduling for sequential printing is also NP-hard, since it is a more general problem.


Genetic optimization has been used for packing 3D objects \cite{Ikonen1997AGA}. Various special algorithms have also been developed for box packing \cite{DBLP:conf/ijcai/LimY01} or algorithms for placing smaller objects in concave parts of other objects \cite{DBLP:journals/eor/EgebladNO07}. The most relevant to us are works that translate problems, whether rectangle packing or 3D packing, into other formalisms, such as those for {\em constraint programming}  (CSP) \cite{DBLP:books/daglib/0016622}, {\em linear programming} LP \cite{rader2010deterministic}, or {\em satisfiability modulo theories} (SMT) \cite{DBLP:reference/mc/BarrettT18} for which an off-the-shelf efficient solvers exist. Rectangle packing using CSP is described in \cite{DBLP:conf/aips/MoffittP06,DBLP:journals/anor/KorfMP10} and the application of SMT for the same problem is shown in \cite{Nikken2020}.





\subsection{Contribution}


We build on top of the previous formalization of the object arranging and scheduling problem for sequential printing as suggested in \cite{DBLP:conf/case/SurynekBMK25,DBLP:conf/iros/SurynekBMK25}, where the problem is denoted SEQ-PACK+S. We explicitly parametrize the previous CEGAR-SEQ with object arrangement and object ordering strategies that we call a {\em composite strategy} - the resulting framework is called Portfolio-CEGAR-SEQ. We also suggest several object orderings and arrangement strategies that are combined together (similarly as a Cartesian product) to form up to 20 composite strategies, a rich {\bf portfolio} but still manageable by a contemporary CPU. Parallel processing of the portfolio on modern CPUs does not add any significant runtime increase but often leads to better solutions in multiple printing plate setting than the original CEGAR-SEQ. Often Portfolio-CEGAR-SEQ is able to sequentially arrange and schedule a large batch of objects on fewer printing plates as shown in our experiments that represents a significant advantage for a printer operator.

\section{Background}



The problem of {\em 3D sequential object arrangement and scheduling} (inspired by the terminology for rectangle packaging, we will rather call the problem ``object packing and scheduling'' and denote SEQ-PACK+S) is a task of determining the positions and order of 3D objects so that the objects can be printed by the 3D printer in a determined order at the determined positions sequentially, one after the other. Unlike standard 3D printing, where all objects are printed at once in individual slices, in sequential printing, objects are completed individually. That is, when the next object is printed, the previously printed objects are still present on the printing plate. It is therefore necessary to ensure that the print head and other mechanical parts of the 3D printer do not collide with previously printed objects when printing the next object.



While determining the printing order of objects to print is a discrete problem, determining object positions is inherently a continuous problem, which presents specific challenges.


Let $\mathbb{R}^3$ be a three dimensional Euclidean space, a finite set of objects $\mathbb{O} = \{ O_1, O_2, ..., O_k\}$, where each object $O_i$ is a connected set $\{(x_i, y_i, z_i) \in O_i\} \subseteq \mathbb{R}^3$, in addition to this, there is an {\em extruder} object $E$, $\{(x_e, y_e, z_e) \in E\} \subseteq \mathbb{R}^3$ that represents a moving part of the printer that prints objects. On a real Cartesian 3D printer, the extruder object is represented by the extruder, print head, gantry, cables, and other moving parts. All objects including the extruder are expressed in the same coordinate system. We assume that the extruder does not change shape during its motion. For simplicity, we abstract from bending of cables in this mathematical model. However, the model can be further generalized to take this into account. Moving extruder $E$ means to translate $E$ to some position $(x_t,y_t,z_t) \in \mathbb{R}^3$ at time $t \in \mathbb{R}$. Hence extruder appears as a translated object $\{(x_e + x_t, y_e + y_t, z_e + z_t) \;|\; (x_e, y_e, z_e) \in E\}\}$ at time $t$. We can assume that $(x_t,y_t,z_t)$ changes smoothly as with real 3D printers happens but it is not important for further definitions.


We can mathematically model the plate as a subset of a plane perpendicular to the $z$-axis. Typically, the plate is rectangular, sometimes circular, exceptionally of a different shape. Mathematically, the plate will be a subset of 2D plane, $P_P \subseteq \mathbb{R}^2$. The extruder moves above the plate using printer mechanics. Any point relatively above the printing plate $P_P$ up to certain height is accessible by the extruder which together defines a {\em printing volume} $\mathbb{V} \subseteq \mathbb{R}^3$.

For simplicity, we will assume that all objects to be printed will fit into the printing volume $\mathbb{V}$, and we will also assume that the extruder can move slightly above the printing volume, which eliminates the need to worry about the height of objects in further definitions.

Common 3D printer mechanics such as {\em Cartesian} (bed-slinger) or CoreXY are covered by our definitions.



From the mathematical point of view, we can look at printing an object in such a way that it is necessary to create every point of the printed object, that is, to touch every point of the printed object with the extruder, or more precisely, with a selected point of the extruder, in our case the point (0,0,0) of the extruder. The point (0,0,0) of the extruder will correspond to the nozzle opening.



The extruder must therefore move to each point of the printed object. Once a point is printed, we must take into account its presence for future movements of the extruder, that is, the extruder must not collide with the point in the future. Mathematically, the printed point must never appear inside the translated extruder.

In standard printing, we assume that we place objects on the plate and print them according to the increasing $z$-coordinate in slices. Assuming that the $z$-coordinates of the extruder are non-negative except for the point (0,0,0) (which also corresponds to a real 3D printer), there is no risk of collision with any already printed part. However, this is no longer case in sequential 3D printing, when individual objects are completed one after other.


In the case of sequential printing, we want to determine the positions of the objects and their temporal ordering, so that when printing objects sequentially in this specified order, there is never a collision between the extruder and a previously printed object. We will call this requirement a sequential {\em non-colliding requirement}. Next, we need to ensure that all objects are placed on the printing plate, we will call this requirement the {\em printing plate requirement} (no part of an object extends beyond the printing plate) \footnote{Equivalently, the printing plate requirement can be expressed by the requirement that all objects are placed within the printing volume $V$. The way of expression we have chosen corresponds better with the proposed solution technique.}. And finally, we need to ensure that the extruder can be lifted vertically up and moved to print the next object, i.e. the extruder can be lowered back to the printing plate for printing the next object. We can simplify the last requirement slightly by requiring the extruder to be able to move freely vertically upwards from any top point of the object being printed. This allows both for initial accessing the volume for printing the object as well as leaving the object volume. We will call this requirement an {\em extruder traversability} requirement. Other variations of this requirement are also possible, and the method we developed is general enough to work with such generalizations.




Formally, we need to determine for each object $O_i \in \mathbb{O}$ its position on the plate $(X_i,Y_i,Z_i) \in \mathbb{R}^3$ ($Z_i$ is determined by placing object vertically on the surface of the plate) and the order of the objects, i.e. the permutation $\pi: \{1,2,...,k\} \rightarrow \{1,2,...,k\}$, such that the {\em sequential non-colliding requirement}, the {\em printing plate requirement} hold, and {\em extruder traversability} hold.

We introduce object transforming functions: $\mathcal{P}: 2^{\mathbb{R}^3} \rightarrow 2^{\mathbb{R}^3}$, a function that places object onto the plate, $\mathcal{E}: 2^{\mathbb{R}^3} \rightarrow 2^{\mathbb{R}^3}$, a function that makes an {\em extruder envelope} of a given object. In addition to this, let $\mathit{()}^{xy}: 2^{\mathbb{R}^3} \rightarrow 2^{\mathbb{R}^2}$ be a {\em projection} of the given object onto the printing plate, and $\mathit{()}^{\top}: 2^{\mathbb{R}^3} \rightarrow 2^{\mathbb{R}^3}$ be an {\em extended top} of the printed object defined as follows:



\begin{itemize}
\item $\mathcal{P}(O_i) = \{(x_i + X_i, y_i + Y_i, z_i + Z_i)\;|\\
\;(x_i,y_i,z_i) \in O_i\}$

\item $\mathcal{E}(O_i) = \{(x_i + x_e, y_i + y_e, z_i + z_e)\;|\\
\;(x_i,y_i,z_i) \in O_i \wedge (x_e,y_e,z_e) \in E\}$


\item $\mathit{O}_i^{xy} = \{(x,y)\;|\;(x,y,z) \in O_i\}$

\item $\mathit{O_i}^{\top} = \{(x,y,z)\;|\;z \geq {\max}_z \wedge (x,y,{\max}_z) \in O_i \}$ \\where ${\max}_z=\max\{z\;|\;(x,y,z) \in O_i\}$
\end{itemize}

Let us note that $\mathcal{E}(\mathcal{P}(O_i))$ and $\mathcal{E}(O_i)$ are equivalent to the Minkowski sum of $\mathcal{P}(O_i)$ and the extruder $E$ and $O_i$ and the extruder $E$ respectively.

The {\bf sequential non-colliding} requirement can be expressed as follows:
\vspace{-0.25cm}

\begin{equation}
\label{con:sequential-noncolliding}
(\forall i,j=1,2,...,k) (\pi(i) < \pi(j) \Rightarrow \mathcal{P}(O_i) \cap \mathcal{E}(\mathcal{P}(O_j)) = \emptyset)
\end{equation}

Placement objects within the printing plate, i.e. the {\bf printing plate} requirement can be expressed as follows:
\vspace{-0.25cm}

\begin{equation}
\label{con:on-plate}
(\forall i=1,2,...,k) (\mathcal{P}(O_i)^{xy} \subseteq P_P)
\end{equation}

The {\bf extruder traversability} requirement can be expressed as follows:
\vspace{-0.25cm}

\begin{equation}
\label{con:extruder-traversability}
(\forall i,j=1,2,...,k) (\pi(i) < \pi(j) \Rightarrow \mathcal{P}(O_i) \cap \mathcal{P}(O_j)^{\top} = \emptyset)
\end{equation}

{\bf Optimality in SEQ-PACK+S}. The problem of object packing for sequential printing is a decision problem in its basic variant. Various notions of optimality in SEQ-PACK+S can be adopted.


The printing plate of a real 3D printer is usually heated to achieve adhesion of objects to the plate, with the most uniform heating towards the center of the plate, while irregularities in heating increase towards the edges of the plate. Given these physical properties, it is advantageous to place printed objects towards the center of the printing plate. This preference has been taken into account in the design of the objective in the original CEGAR-SEQ algorithm. Let $C_{P} \in \mathbb{R}^2$ be a center of $P_P$.

Intuitively, we will try to shrink $P_P$ around  $C_{P}$ so that placement of object satisfying the sequential printability requirements is still possible. Formally, let $C_{P}=(x_c,y_c)$ and $\sigma{P_P} = \{(x_c+\sigma(x - x_c), y_c+\sigma(y - y_c))\;|\;(x,y) \in P_P\}$. The printing plate requirement can be modified accordingly:

\begin{equation}
\label{con:on-plate-opt}
(\forall i=1,2,...,k) (\mathcal{P}(O_i)^{xy} \subseteq \sigma{P_P})
\end{equation}

We will define the optimization variant of SEQ-PACK+S as finding $\sigma \in (0,1]$ that is as small as possible and requirements \ref{con:sequential-noncolliding}, \ref{con:extruder-traversability}, and \ref{con:on-plate-opt} are satisfied.

\section{A Linear Arithmetic Model}

Let $X_i, Y_i \in \mathbb{R}$ be decision variables determining the position of object $O_i$ and let $T_i \in \mathbb{R}$ be decision variables determining the time at which respective objects $O_i$ are printed. $T_i$ variables will be used to determine the permutation $\pi$ of objects for sequential printing.

A linear arithmetic formula that expresses the SEQ-PACK+S problem over $X_i, Y_i$ and $T_i$ is constructed. For simplicity we recall here only important blocks of constraints, for details of the formula we refer the reader to \cite{DBLP:conf/iros/SurynekBMK25}.

Let $P_A=(A_1,A_2,...A_{\alpha}) \subseteq \mathbb{R}^2$ and $P_B=(B_1,B_2,...,B_{\beta}) \subseteq \mathbb{R}^2$  are two polygons. A constraint {\em Points-outside-Polygon} or $\mathit{PoP}(X_A, Y_A, P_A, X_B, Y_B, P_B)$ requires that vertices of $P_A$ placed at position $X_A,Y_A$ are outside of $P_B$ placed at position $X_A,X_B$. This constraint is relatively easy to express, however is not sufficient to express the requirement that polygons do not overlap.

To ensure non-overlapping between polygons a {\em Lines-not-Intersect} or $\mathit{LnI}(X_A, Y_A, A_i,A_{(i\bmod k)+1},$ $X_B,Y_B,B_j,B_{(j \bmod l)+1})$ constraint is needed. The constraint ensures that the $i$-th edge of polygon $A$ placed at position $X_A,Y_A$ does not intersect with the $j$-th edge of polygon $B$ placed at position $X_B, Y_B$. Since this constraint is harder to express, the CEGAR-SEQ algorithm does not add these constraints from the beginning. These constraints are rather treated in the counterexample guided abstraction refinement style (CEGAR) \cite{DBLP:conf/cav/ClarkeGJLV00,DBLP:journals/jacm/ClarkeGJLV03}. Briefly said, the $\mathit{LnI}$ constraint is added only after if it is violated. As shown in \cite{DBLP:conf/iros/SurynekBMK25} the CEGAR-style constraint refinement is a key to efficiency of the CEGAR-SEQ algorithm.

For expressing the requirement that a polygon lies inside the printing plate a constraint {\em Polygon-inside-Polygon} or $\mathit{PiP}(X_A, Y_A, P_A, X_B, Y_B,P_B)$ is used. The constraint expresses that vertices of $P_A$ placed at position $X_A,Y_A$ are inside $P_B$ placed at position $X_B,Y_B$.

The CEGAR-SEQ algorithm for solving an instance of the approximation of SEQ-PACK+S as a linear arithmetic formula by a CEGAR-inspired approach is shown using pseudo-code as Algorithm \ref{alg:parametrized-CEGAR-SEQ}. The pseudo-code shows a variant of the algorithm that is parametrized by a composite strategy. Let us briefly recall that the algorithm introduces the constraints into a formula $\mathcal{F}$ that is subsequently solved by an SMT solver. For details on the CEGAR-SEQ algorithm we refer the reader to \cite{DBLP:conf/iros/SurynekBMK25}.

\section{A Portfolio of Scheduling Strategies}

\begin{figure}[h]
    \centering
    \includegraphics[trim={3.0cm 23.0cm 3.1cm 3.0cm},clip,width=0.75\textwidth]{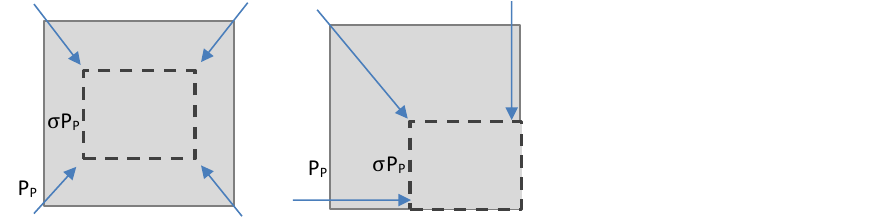}
    \vspace{-0.4cm}
    \caption{Object arrangement strategies (tactics): (i) objects are placed towards the center of the printing plate $P_P$ ({\bf Center} tactic), (ii) objects are placed towards the maximum $x$ and minimum $y$ coordinates w.r.t. $P_P$ ({\bf Max-X-Min-Y} tactic).}
    \label{fig:tactics}
    \vspace{-0.5cm}        
\end{figure}

The original CEGAR-SEQ algorithm uses an object arrangement strategy that places objects towards the center of the printing plate. This is justified by the fact that heat distribution over the printing plate is more regular towards its center (see Figure \ref{fig:tactics} - left). However modern 3D printers are often enclosed and advancements in the design of heated beds eliminate irregularities in the heat distribution. Therefore it starts to be worthwhile to consider different object arrangement and scheduling strategies such as arranging objects towards a corner of the printing plate (see Figure \ref{fig:tactics} - right).

\begin{algorithm}[t!]
\begin{footnotesize}
\SetKwBlock{NRICL}{Solve-CEGAR-SEQ$(P_P,\mathbb{O},\mathit{STRATEGY})$}{end} \NRICL{
    $\mathit{Decision} \gets []$\\
    \While {$\mathbb{O} \neq \emptyset$} {
        $\{O_1,O_2,...,O_k\} \gets \mathit{STRATEGY}$.Ordering($\mathbb{O}$)\\
        $(X_1,Y_1,T_1),...,(X_k,Y_k,T_k)) \gets ((\bot,\bot,\bot),...,(\bot,\bot,\bot))$\\            
        
       $\mathcal{F} \gets []$ \\
       \For{each $i,j \in \{1,2,...,k\} \wedge i \neq j$}{
           $\mathcal{F} \gets \mathcal{F} \cup \{T_i  + \epsilon_T < T_j \vee T_j  + \epsilon_T < T_i\}$\\
           $\mathcal{F} \gets \mathcal{F} \cup \{T_i < T_j \Rightarrow \; (\mathit{PoP}(X_i, Y_i, \mathcal{C}(O_i^{xy}),
                                                                                                               X_j, Y_j, \mathcal{C}(\mathcal{E}(O_j)^{xy})) 
                                                           \wedge \; \mathit{PoP}( X_j, Y_j, \mathcal{C}(\mathcal{E}(O_j)^{xy}),
                                                                                        X_i, Y_i, \mathcal{C}(O_i^{xy})))\}$
        }
        $\sigma_{0} \gets 0$\\
        $\sigma_{+} \gets 1$\\    
        \While {$\sigma_{+} - \sigma_{0} > \epsilon_{XY}$} {
      	    $\sigma \gets (\sigma_{+} + \sigma_{0}) / 2$ \\
      	    ${\sigma}P_P \gets  \mathit{STRATEGY}$.Tactic($P_P, \sigma$)\\
    	    $\mathit{answer} \gets$ Solve-CEGAR-SEQ-Bounded ($\mathcal{F},{\sigma}P_P,\{O_1,O_2,...,O_k\})$\\
    	    \If {$\mathit{answer} \neq \mathit{UNSAT}$}{
    		  {\bf let} $(((X_1,Y_1,T_1),...,$\\$(X_k,Y_k,T_k)),\mathcal{F}) = \mathit{answer}$\\
    	    	  $\sigma_{+} \gets \sigma$
    	    }
    	    \Else{
    		$\sigma_{0} \gets \sigma$
    	     }
         }
         $\mathbb{O} \gets \mathbb{O} \setminus \{O_1,O_2,...,O_k\}$\\
         $\mathit{Decision} \gets \mathit{Decision}.((X_1,Y_1),...,(X_k,Y_k))$\\
     }
     \Return{$\mathit{Decision}$}
}

\SetKwBlock{NRICL}{Solve-CEGAR-SEQ-Bounded($\mathcal{F},{\sigma}P_P,\{O_1,O_2,...,O_k\}$)}{end} \NRICL{
	$\Phi \gets []$\\
	\For{each $i \in \{1,2,...,k\}$}{
	    $\Phi \gets \Phi \cup \{\mathit{PiP}(X_i,Y_i,\mathcal{C}(O_i^{xy}),0,0,\sigma P_P)\}$
	}
	\While{$\mathit{TRUE}$}{
 	  $\mathit{answer} \gets$ Solve-SMT$(\mathcal{F},\Phi)$\\	
	  \If{$\mathit{answer} \neq \mathit{UNSAT}$}{
	    {\bf let} $((X_1,Y_1,T_1),...,(X_k,Y_k,T_k)) = \mathit{answer}$\\
	    \For{each $i,j \in \{1,2,...,k\} \wedge i \neq j$} {        
		{\bf let} $(A_1,A_2,...,A_{\alpha}) = \mathcal{C}(O_i^{xy})$ \\
		{\bf let} $(B_1,B_2,...,B_{\beta}) =  \mathcal{E}(\mathcal{C}(O_j^{xy}))$ \\
		$\mathit{refined} \gets \mathit{FALSE}$\\
		\For{each $a \in \{1,2,...,\alpha\}$}{
		  	\For{each $b \in \{1,2,...,\beta\}$}{
		  	    \If{$T_i < T_j \wedge$ Lines-Intersect$(X_i, Y_i, A_a,$\\
		  	                            $A_{(a\bmod \alpha)+1},
		  	                             X_j, Y_j,B_b,B_{(b\bmod \beta)+1})$}{
                           $\mathcal{F} \gets \mathcal{F} \cup \{T_i < T_j \Rightarrow
                                                                                                 \mathit{LnI}(X_i, Y_i, A_a,A_{(a\bmod \alpha)+1},$ \\
                                                                                                 $X_j, Y_j,B_b,B_{(b\bmod \beta)+1})\}$\\
                           $\mathit{refined} \gets \mathit{TRUE}$\\
		  	    }
		  	}
		}
		\If{$\neg \mathit{refined}$}{
		    \Return {$(((X_1,Y_1,T_1),...,(X_k,Y_k,T_k)),\mathcal{F})$}		
		}
	    }
       }
       \Else{
         \Return{$\mathit{UNSAT}$}
       }
    }
}

\caption{CEGAR-SEQ \cite{DBLP:conf/iros/SurynekBMK25} parameterized by a composite strategy: sequential object packing by a CEGAR-based algorithm where a given strategy determines object ordering and object arrangement tactic.}
\label{alg:parametrized-CEGAR-SEQ}
\end{footnotesize}
\end{algorithm}

In addition to object arrangement strategies it may be beneficial to consider various object orderings in the context of sequential arranging and scheduling objects across multiple printing plates. Object ordering determines which objects will be arranged and scheduled for the first plate, which objects will go to the second plate, etc. More precisely, given object ordering, the algorithm will try to arrange and schedule the first $k$ objects according to the ordering for the first printing plate where $k$ is as large as possible.

We denote spatial object arrangement strategy on plate as {\em tactic} and object ordering simply as {\em ordering}. A {\em composite strategy} denoted as $\mathit{STRATEGY}$ for short consists of:
\begin{itemize}
  \item an {\bf object arrangement} strategy, that is a tactic, denoted $\mathit{STRATEGY}.\mathit{Tactic}$ and
  \item an {\bf object ordering}, denoted $\mathit{STRATEGY}.\mathit{Ordering}$.
 \end{itemize}
 
The original CEGAR-SEQ algorithm has been modified so that it takes the composite strategy $\mathit{STRATEGY}$ as its input. The modified CEGAR-SEQ combines a tactic and an ordering from the given composite strategy for selection of objects for arrangement on the plate and for the way in which objects are arranged on the plate.

The following tactics are used to compose composite strategies, the set denoted $\mathit{Tactics}$:

\begin{itemize}
  \item {\bf Center} - objects are arranged towards the center of the plate (see Figure \ref{fig:tactics} - left)
  \item{\bf Max-X-Min-Y} - object are arranged towards the maximum $x$ and the minimum $y$ coordinates of the plate (see Figure \ref{fig:tactics} - right)
  \item {\bf Min-X-Max-Y}, {\bf Min-X-Min-Y}, and {\bf Max-X-Max-Y} - with analogous meaning
\end{itemize}

The abstract model of a given tactic is carried out via a function $\mathit{STRATEGY}.\mathit{Tactic}(\sigma, P_P)$ that for a given $\sigma \in (0,1]$ and a printing plate $P_P$ returns the printing plate scaled down by a factor of $\sigma$ while the resulting printing plate denoted $\sigma P_P$ is shaped according to the given tactic (towards the center or towards a corner, etc.).

The abstract model of an ordering is for simplicity carried out by a non-deterministic function $\mathit{STRATEGY}.\mathit{Ordering}(\mathbb{O})$ that for a given (finite) set of objects $\mathbb{O}$ returns a subset of objects $\{O_1, O_2, ..., O_k \} \subseteq \mathbb{O}$ that fits with respect to sequential printing onto the plate, moreover $k$ is as large as possible, that is, the next object $O_{k+1}$ with respect to the ordering does not fit on the plate (objects $\{O_1, O_2, ..., O_{k+1} \}$ cannot be arranged and scheduled sequentially on the plate). The abovementioned non-determinism allows for easy guessing of $k$ \footnote{Without the help of non-determinism we would have to add another optimization loop to the algorithm for finding optimal $k$ which would make the algorithms' pseudo-code too complicated.}.

Suggested object orderings is based on expert knowledge that the height of the object is a decisive property for sequential printing. Therefore suggested object orderings sorts objects according to their height. The following object orderings are used to compose composite strategies, the set denoted $\mathit{Orderings}$:

\begin{itemize}
\item {\bf Height-Min-to-Max} - objects are ordered according to their height from the shortest one to the tallest one
\item {\bf Height-Max-to-Min} - objects are ordered according to their height from the tallest one to the shortest one
\item {\bf Height-Random} - objects are ordered randomly
\item {\bf Height-Input} - objects are taken in the same order as they come in the input
\end{itemize}

Altogether the basic portfolio of the for the modified CEGAR-SEQ algorithm consists of $|\mathit{Tactics} \times \mathit{Orderings}|$ composite strategies. \footnote{Concretely it is $5 \times 4 = 20$ which is adequate for parallel processing by a multi-core CPU of a contemporary personal computer.}. The pseudo-code of Portfolio-CEGAR-SEQ is shown as Algorithm \ref{alg:portfolio-CEGAR-SEQ}.

\begin{algorithm}[t!]
\begin{footnotesize}
\SetKwBlock{NRICL}{Portfolio-CEGAR-SEQ$(P_P,\mathbb{O})$}{end} \NRICL{
    $\mathit{Orderings} \gets \{ \mathit{Height\text{-}Min\text{-}to\text{-}Max},$\\$
    \;\;\;\;\;\;\;\;\;\;\;\;\;\;\;\;\;\;\;\;\;\;\;\;\mathit{Height\text{-}Max\text{-}to\text{-}Min}, \mathit{Height\text{-}Random} \}$\\
    $\mathit{Tactics} \gets \{ \mathit{Center}, \mathit{Min\text{-}X\text{-}Min\text{-}Y},\mathit{Max\text{-}X\text{-}Min\text{-}Y},$\\
    						$\;\;\;\;\;\;\;\;\;\;\;\;\;\;\;\;\;\;\;\;\mathit{Min\text{-}X\text{-}Max\text{-}Y}, \mathit{Max\text{-}X\text{-}Max\text{-}Y} \}$\\
   $\mathit{Answers} \gets \emptyset$\\
    \For {$\mathit{ordering} \in \mathit{Orderings}$} {
	    \For {$\mathit{tactic} \in \mathit{Tactics}$} {
    	        $\mathit{STRATEGY}$.Ordering$ \gets \mathit{ordering}$\\
    	        $\mathit{STRATEGY}$.Tactic$ \gets \mathit{tactic}$\\
    	        $\mathit{answer} \gets$ Solve-CEGAR-SEQ$(P_P, \mathbb{O},\mathit{STRATEGY})$\\
    	        $\mathit{Answers} \gets \mathit{Answers} \cup \{answer\}$\\
	    }
     }
     $\mathit{best}\text{-}\mathit{answer} \gets$ Select-Best-Schedule$(\mathit{Answers})$\\
     \Return{$\mathit{best}\text{-}\mathit{answer}$}
}

\caption{Portfolio-CEGAR-SEQ: A portfolio-based high level extension of the CEGAR-SEQ algorithm - object ordering and arrangement tactics are combined. For loops at line 7 and 8 are expected to be implemented in {\bf parallel}.}
\label{alg:portfolio-CEGAR-SEQ}
\end{footnotesize}
\end{algorithm}

\begin{figure*}[t]
    \centering
    \begin{subfigure}{0.49\textwidth}
       \centering
       \includegraphics[trim={2.5cm 22.0cm 10.5cm 2.5cm},clip,width=0.95\textwidth]{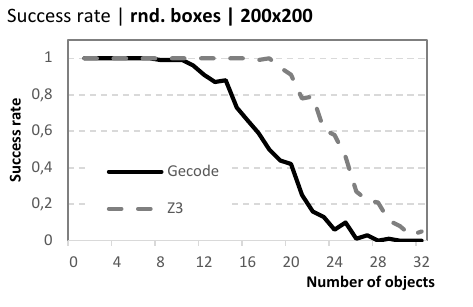}
    \end{subfigure}
    \begin{subfigure}{0.49\textwidth}
       \centering
       \includegraphics[trim={2.5cm 22.0cm 10.5cm 2.5cm},clip,width=0.95\textwidth]{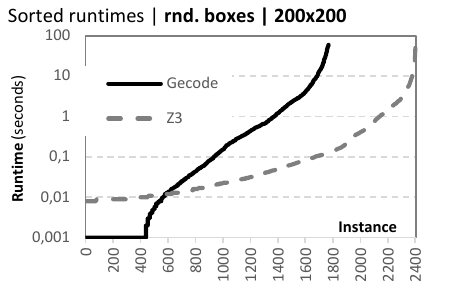}
    \end{subfigure}
    \caption{Comparison of solving SEQ-PACK+S by the Gecode solver and the z3 solver on random cuboids on a 200mm $\times$ 200mm plate. The right part shows cactus plots of runtimes (lower plot is better).}
    \label{expr:solvers-200x200}
    \vspace{-0.5cm}    
\end{figure*}

\section{Experimental Evaluation}

CEGAR-SEQ \cite{DBLP:conf/iros/SurynekBMK25} has been written in C++ and integrated as part of Prusa Slicer 2.9.1 \cite{prusa-slicer-2025}, an open-source slicing software for 3D printers. Portfolio-CEGAR-SEQ has been written in C++ as well but currently it is not integrated within the publicly available slicing software.

Portfolio-CEGAR-SEQ shares with the previous CEGAR-SEQ the internal usage of the Z3 Theorem Prover \cite{10.5555/1792734.1792766}, an SMT solver, that has been used for solving the linear arithmetic model. We have performed extensive testing of both the CEGAR-SEQ algorithm and Portfolio-CEGAR-SEQ and present some of the results in this section. All experiments were performed on a system with CPU AMD Ryzen 7 2700 3.2GHz, 32GB RAM, running Kubuntu Linux 24.

All source code and experimental data presented in this paper are available in \texttt{github.com/\{double-blind\}/cegar-seq} repository to support reproducibility of results.


\subsection{Comparison of Underlying Solvers}


\begin{figure}[t]
    \centering
    \includegraphics[trim={2.5cm 22.0cm 10.5cm 2.5cm},clip,width=0.49\textwidth]{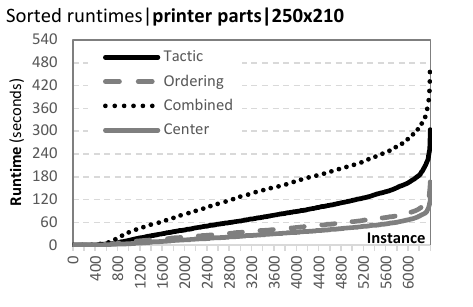}
    \caption{Sorted runtimes of Portfolio-CEGAR-SEQ with various portfolios on printer parts.}
    \label{expr:portfolio-runtimes}
    \vspace{-0.5cm}        
\end{figure}

Since there is currently no comparable other comparable solver for sequential printing other than CEGAR-SEQ and Portfolio-CEGAR-SEQ, we decided to conduct a competitive comparison at least at the solver level.

\begin{figure*}[t]
    \centering
    \begin{subfigure}{0.49\textwidth}
       \centering
       \includegraphics[trim={2.5cm 22.0cm 10.5cm 2.5cm},clip,width=0.95\textwidth]{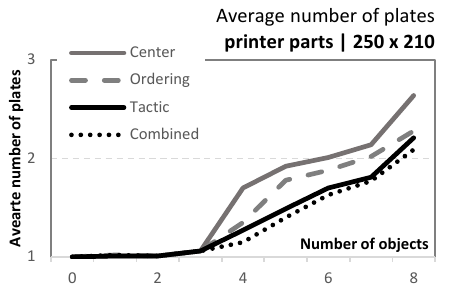}
    \end{subfigure}
    \begin{subfigure}{0.49\textwidth}
       \centering
       \includegraphics[trim={2.5cm 22.0cm 10.5cm 2.5cm},clip,width=0.95\textwidth]{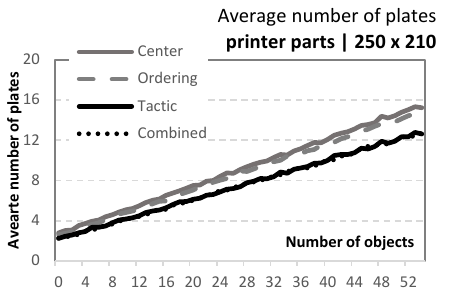}
    \end{subfigure}
    \caption{Average number of printing plates used by a sub-optimal variant of Portfolio-CEGAR-SEQ under four different portfolios of composite strategies.}
    \label{expr:portfolio-plates}
\end{figure*}

We compared solving of the SEQ-PACK+S problem with the Z3 solver and with the Gecode solver \cite{gecode2006}, which is a CSP solver. Hence we compared solving SEQ-PACK+S in two different paradigms, SMT and CSP.

The experimental setup consisted of a printing plate of size 200 $\times$ 200 \footnote{Sizes can be assumed in mm that corresponds to a real 3D printer.} and random cuboids whose length, width and height were integer and chosen randomly from a uniform distribution in the interval [8,64], the number of cuboids ranges from 1 to 32. For each number of cuboids, 100 random instances were generated. The timeout was set to one minute. This experiment reproduces the experiment from  \cite{DBLP:conf/iros/SurynekBMK25} but adds more measurements per data point (100 vs. 10 in the original experiment) and provides greater timeout (60 seconds vs. 8 seconds in the original experiment).

The solvers in this test were set to an optimal mode which means that all objects were scheduled at once into a as small as possible rectangle placed around the center of the printing plate.


Results are shown in Figure \ref{expr:solvers-200x200} which shows a clear dominance of the SMT paradigm represented by the Z3 solver. Additionally, we note that the linear arithmetic formula uses variables with a rational domain in contrast to finite domain of Gecode with millimeter resolution, so the resulting solution from the Z3 solver is more accurate.

In the optimal mode, there is a certain limit for the number of objects the solvers are able to solve within the given timeout of 60 seconds. Results indicate that scheduling more than 25 cuboids is virtually unsolvable for the Gecode solver; this limit is approximately 30 cuboids for the Z3 solver.

We also performed runtime comparison between various variants of the Portfolio-CEGAR-SEQ solver and the original CEGAR-SEQ. We used 4 setups of composite strategies:

\begin{itemize}
  \item {\bf Center} - corresponds to the original CEGAR-SEQ with a single {\em Center} tactic combined with {\em Height-Input} object ordering
  \item {\bf Ordering}  - consists of four object orderings {\em Height-Min-to-Max}, {\em Height-Max-to-Min}, {\em Height-Random}, and {\em Height-Input} combined with the {\em Center} tactic
  \item {\bf Tactic} - consists of five tactics {\em Center}, {\em Min-X-Min-Y}, {\em Max-X-Min-Y}, {\em Min-X-Max-Y}, and {\em Max-X-Max-Y} cobined with {\em Height-Input} object ordering
  \item {\bf Combined} - consists of all object orderings and tactics from previous two composite strategies that altogether form 20 composite strategies
\end{itemize}
 
This experiment has a different setup. As a testing case we use sequential 3D printing of complex objects - 3D printable parts for a 3D printer \footnote{The benchmark consists of 3D printable parts for the Original Prusa MK3S printer.}, the benchmark consists of 34 diverse objects. Similar setup has been used in \cite{DBLP:conf/iros/SurynekBMK25}.

We tried to solve SEQ-PACK+S for the increasing number of objects ranging from 1 up to 64. For each number of objects we selected random objects from the benchmark set (objects can be repeated), 100 instances per number of objects were solved.

Portfolio-CEGAR-SEQ has been set to a sub-optimal mode in this experiment. This is achieved through the $\mathit{STRATEGY}.\mathit{Ordering}$ that returns at most $k$ objects that are arranged and scheduled with respect to previously arranged and scheduled objects. Each time given $k$ objects are scheduled optimally. Since objects are scheduled in groups of $k$ the algorithm cannot reach global optimum, but the advantage is that it is much faster. Parameter $k$ was set to 4 in this experiment. In addition to this,  solving takes place across multiple printing plates - if objects do not fit onto a plate then remaining objects are scheduled on a fresh printing plate. $\mathit{STRATEGY}. \mathit{Ordering}$ is assumed to non-deterministically guess and return the remaining number of objects for the plate if it less than $k$.

Portfolio-CEGAR-SEQ in the sub-optimal mode cannot normally fail to arrange and schedule even many objects as it can use multiple printing plates if needed while only a bounded number of objects is arranged and scheduled on a single plate. Runtime results are shown in Figure \ref{expr:portfolio-runtimes}. The algorithm has been run in parallel for all composite strategies in a given setup and finished after object arranging and scheduling has been finished for all the composite strategies (the best solution is eventually returned).

Results clearly indicate that adding more composite strategies increases wall-clock runtime up to multiple times. However, even for 20 composite strategies the increase factor in wall-clock runtime is acceptable.




\subsection{Benefit of Portfolio}

\begin{figure}[h]
    \centering
    \includegraphics[trim={2.5cm 22.0cm 10.5cm 2.5cm},clip,width=0.45\textwidth]{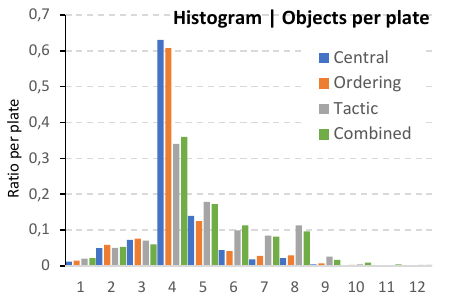}
    \vspace{-0.2cm}
    \caption{Histogram of the number of plates used by a sub-optimal variant of Portfolio-CEGAR-SEQ under four different portfolios of composite strategies.}
    \label{expr:portfolio-histogram}
\end{figure}

\begin{figure*}[t]
    \centering
    \begin{subfigure}{0.31\textwidth}
       \includegraphics[trim={0.5cm 0.5cm 0.5cm 0.5cm},clip,width=1.0\textwidth]{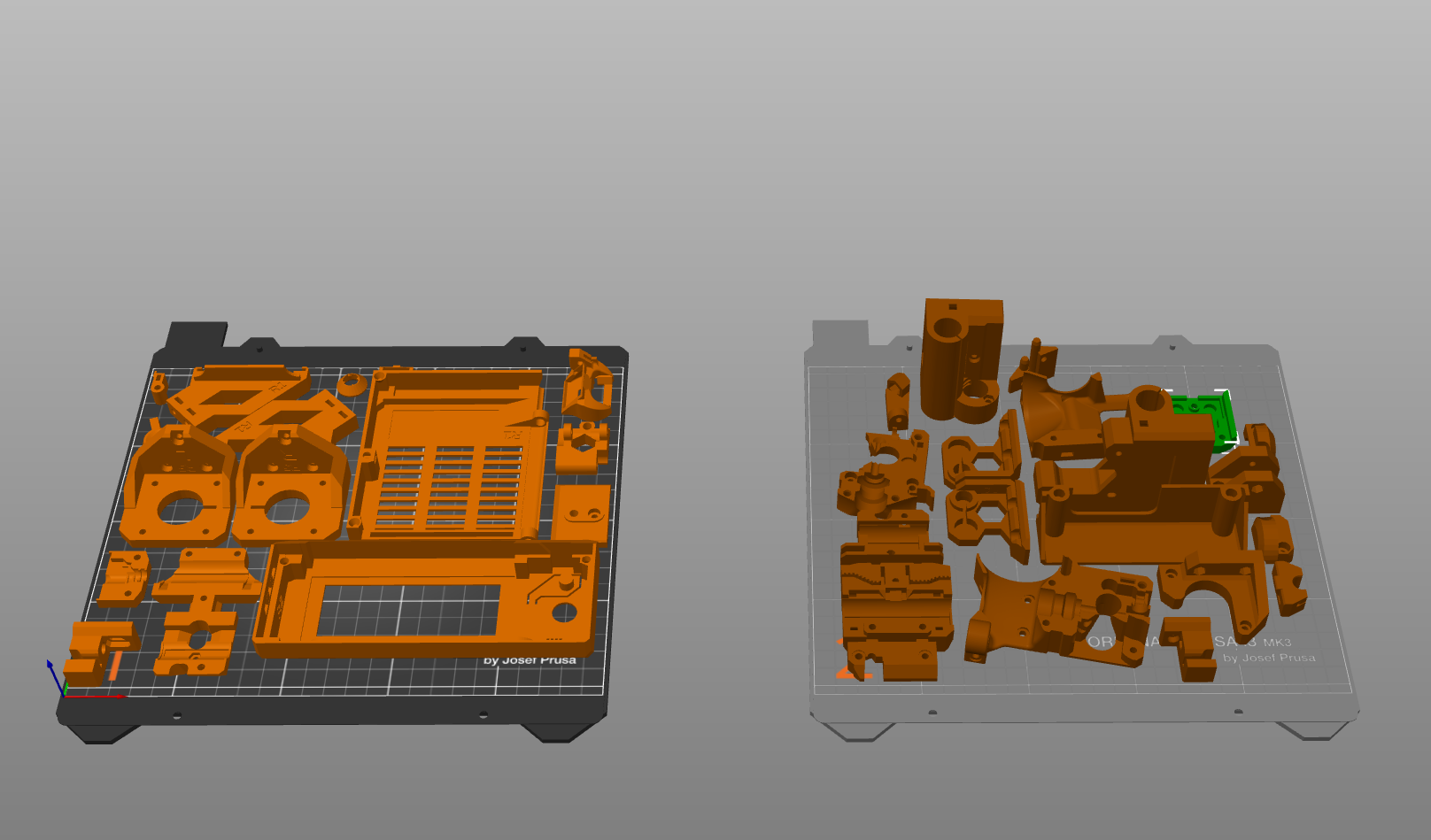}
    \end{subfigure}
    \begin{subfigure}{0.33\textwidth}
       \includegraphics[trim={0.5cm 0.5cm 0.5cm 0.5cm},clip,width=1.0\textwidth]{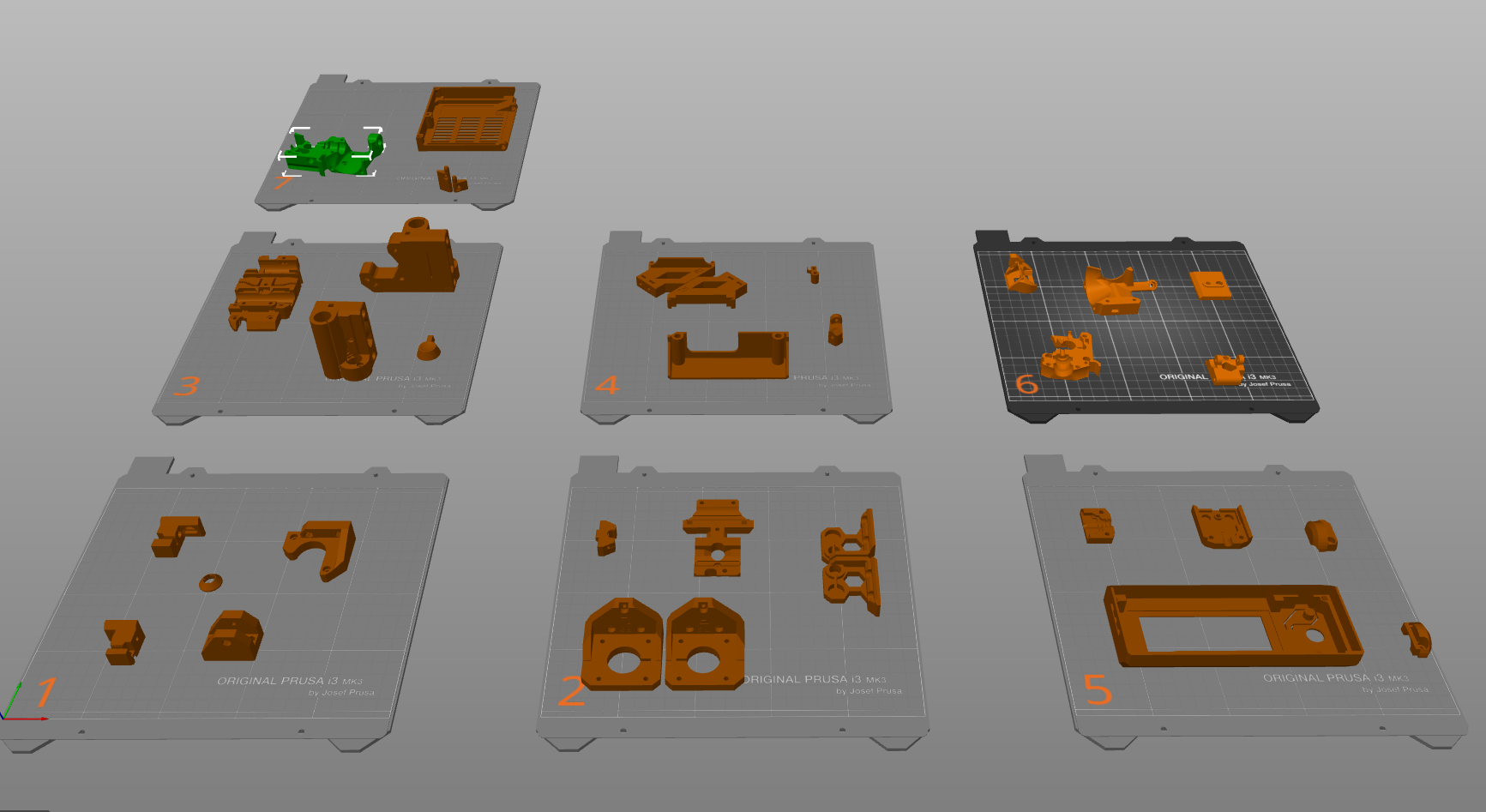}
    \end{subfigure}
    \begin{subfigure}{0.34\textwidth}
       \includegraphics[trim={0.5cm 0.5cm 0.5cm 0.5cm},clip,width=1.0\textwidth]{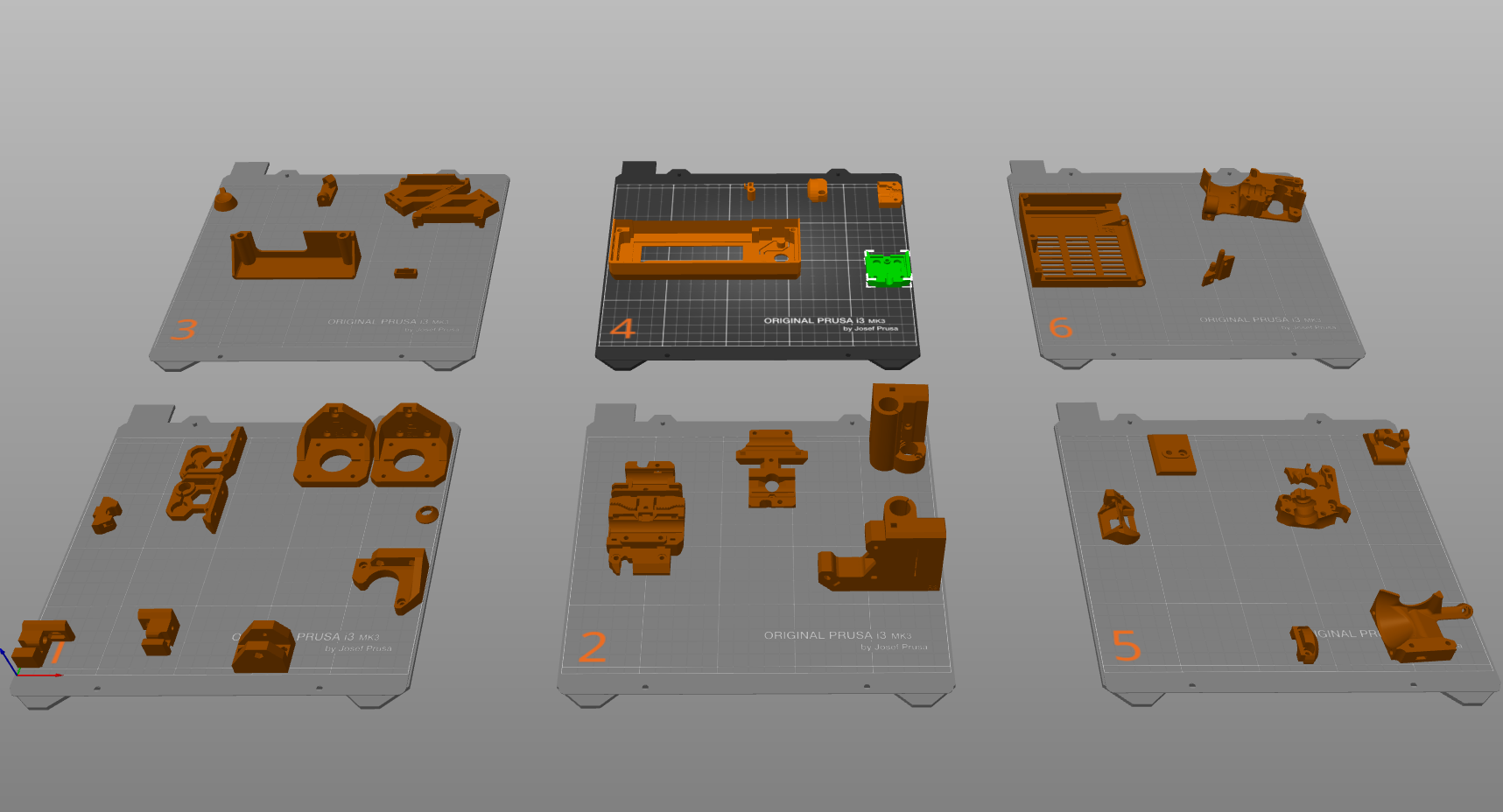}
    \end{subfigure}    
    \caption{The benefit of portfolio-based sequential print scheduling for 30 pieces of printer parts - left: the standard parallel printing. Middle: objects scheduled sequentially by the previous CEGAR-SEQ (7 plates used). Right: objects scheduled sequentially by Portfolio-CEGAR-SEQ (6 plates used).}
    \label{fig:portfolio-benefit}
\end{figure*}

The main benefit of portfolio-based solver for sequential printing is to produce better solutions in the sub-optimal setting. The quality of solutions can be measured by the number of printing plates being used. For example fitting objects onto 1 plate instead of 2 represents a significant practical difference for the operator of the printer.

To evaluate the benefit of using Portfolio-CEGAR-SEQ, we measured the number of printing plates used for arranging and scheduling by the previous CEGAR-SEQ and and the new Portfolio-CEGAR-SEQ in their sub-optimal setup (groups of objects of size 4 are scheduled at once). Results are shown in Figure \ref{expr:portfolio-plates}. For a small number of objects, the richer the portfolio is, the fewer number of plates are being used. A printing plate is often saved, which for a small batch of objects to print represents a significant difference. The best results can be achieved by the {\bf Combined} portfolio. The {\em Ordering} and the {\em Tactic} composite strategies exhibit an orthogonal behavior when their synergy comes to an effect in the {\em Combined} composite strategies. The {\em Tactic} component brings a greater benefit according to experiments.

As the number of objects grows the benefit of the {\em Combined} composite strategies is diminishing with respect to the second best {\em Tactic} composite strategies. Nevertheless, the benefit of Portfolio-CEGAR-SEQ is still significant with respect to the original CEGAR-SEQ - significantly fewer printing plates are often used for scheduling.

The histogram of the number of per printing plate is shown in Figure \ref{expr:portfolio-histogram}. Results indicate that more complex composite strategies tend to place more objects onto the printing plate. The original CEGAR-SEQ and {\em Ordering} composite strategies tend to place $k$ objects onto printing plate in many cases, that is the same number of objects as returned by $\mathit{STRATEGY}.\mathit{Ordering}$. The more complex composite strategies tend to go beyond $k$ objects per printing plate.


\section{Conclusion}

We proposed an extension of CEGAR-SEQ \cite{DBLP:conf/iros/SurynekBMK25}, an existing algorithm for sequential object arrangement and scheduling, with a {\bf portfolio} of object ordering and arrangement strategies. Our new algorithm Portfolio-CEGAR-SEQ runs instances of CEGAR-SEQ with various strategies in parallel and eventually chooses the best solution according to the given objective found by one of the parallel instances. The number of printing plates being used for sequential printing of a large batch of objects turned out to be a useful objective.

Portfolio-CEGAR-SEQ has been shown to generate better solutions in terms of the number of printing plates being used - Portfolio-CEGAR-SEQ often uses fewer printing plates which represents a significant advantage of the printer operator. Moreover, it turned out that object ordering and arrangement strategies have an orthogonal effect to some extent, that is, they can be combined to achieve a synergistic effect as shown in our experiments.

For future work, we want to deal with the possibility of rotating objects directly in the formal model. An open question is also how to integrate heuristics for increasing performance - one option seems to be initialization of the solver with a heuristically found solution.

\bibliographystyle{IEEEtran}
\bibliography{references}

\vfill

\end{document}